%% file: main.tex
\documentclass[10pt,twocolumn,letterpaper]{article}

\usepackage[pagenumbers]{cvpr}

\input{preamble}

\definecolor{cvprblue}{rgb}{0.21,0.49,0.74}
\usepackage[pagebackref,breaklinks,colorlinks,allcolors=cvprblue]{hyperref}

\def\paperID{23} %
\def\confName{CVPR}
\def\confYear{2025}

\title{\textsc{FIction}: 4D Future Interaction Prediction from Video}

\author{Kumar Ashutosh, Georgios Pavlakos, Kristen Grauman\\
University of Texas at Austin
}

\begin{document}
\maketitle
\input{sec/0_abstract}

\input{sec/1_introduction_new}

\input{sec/2_related_work}

\input{sec/3_method}
\input{sec/4_experiments}

\input{sec/5_conclusion}

{
    \small
    \bibliographystyle{ieeenat_fullname}
    \bibliography{main}
}

\input{sec/X_suppl}

\end{document}

%% file: preamble.tex
\usepackage{lipsum}
\usepackage{bbm}
\usepackage{multirow}
\usepackage{array}
\usepackage{booktabs} %
\usepackage{colortbl}
\usepackage{tcolorbox}
\tcbuselibrary{breakable}

\newcommand{\PreserveBackslash}[1]{\let\temp=\\#1\let\\=\temp}
\newcolumntype{C}[1]{>{\PreserveBackslash\centering}p{#1}}
\newcolumntype{L}[1]{>{\PreserveBackslash\raggedright}p{#1}}

\usepackage[accsupp]{axessibility}

\newcommand{\modelname}{\textsc{FIction}}
\definecolor{Gray}{gray}{0.9}

%% file: sec/0_abstract.tex
\begin{abstract}

Anticipating how a person will interact with objects in an environment is essential for activity understanding,
but existing methods are limited to the 2D space of video frames---capturing physically ungrounded predictions of ``what" and ignoring the ``where" and ``how".   
We introduce \modelname~for 
 4D \underline{f}uture \underline{i}nteraction predi\underline{ction} from videos.  
Given an input video of a human activity, the goal is to predict which %
objects at what 3D locations the person will interact with 
in the next time period (e.g., cabinet, fridge), %
and how they will execute that interaction (e.g., poses for bending, reaching, pulling).  %
Our novel model \modelname~fuses the past video observation of the person's actions and their environment to predict both the ``where" and ``how" of future interactions.
Through comprehensive experiments on a variety of activities and real-world environments in Ego-Exo4D, we show that our proposed approach outperforms prior autoregressive and (lifted) 2D video models substantially, with more than 30\% relative gains. 
\renewcommand{\thefootnote}{}\footnotetext{Project Page: \href{https://vision.cs.utexas.edu/projects/FIction/}{https://vision.cs.utexas.edu/projects/FIction/}}

\end{abstract}

%% file: sec/1_introduction_new.tex
\section{Introduction}

\begin{figure*}
    \centering
    \includegraphics[width=\linewidth]{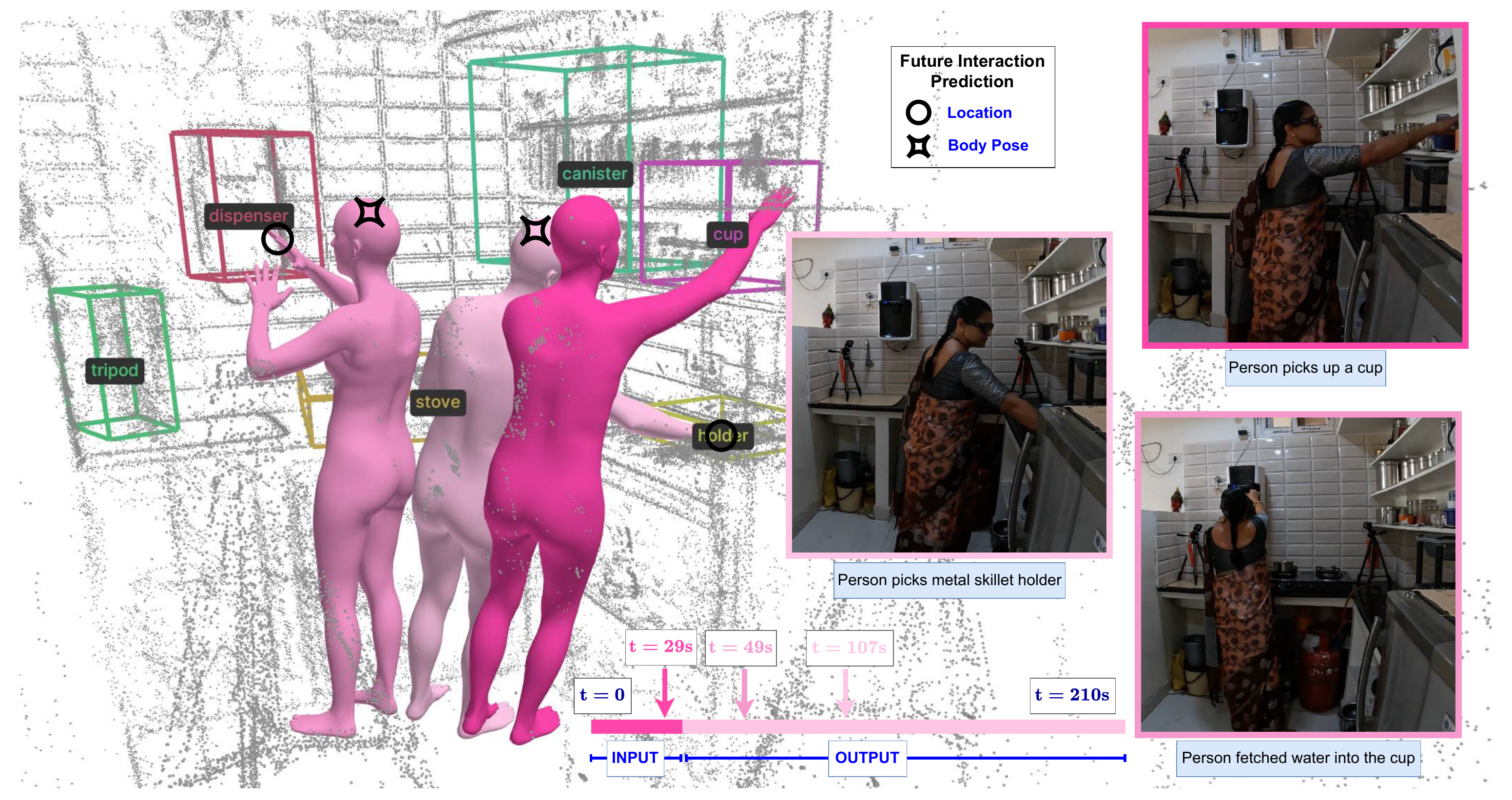}
    \caption{\textbf{Future interaction prediction.}
    When doing a procedure like \emph{making milk tea},
    a person moves around in their environment, interacting with different objects like \emph{water dispenser}, \emph{stove}, \emph{skillet holder}, and \emph{cups}. Each interaction has an associated body pose, e.g., using two hands when fetching water, extending the body to reach upper wall cabinets. Given an environment and the procedure till a time $t$, we predict \emph{all} future object interactions (specifically, pooled over the next 3 mins) and the likely body poses during those object interactions.
    Best viewed in zoom. Only representative object bounding boxes shown for clarity.
    }
    \label{fig:teaser}
    \vspace{-0.5cm}
\end{figure*}

Humans constantly move around their environment and interact with objects to accomplish various daily tasks. A person making a salad will first get a bowl from the cabinet, get lettuce from the fridge, and then dressings from the shelf. An assistive AI agent, by observing her intent, could help her in various ways---fetching a better dressing for her, or helping her get the bowl if she has trouble bending to the lower cabinet. Such an AI assistant needs to understand (a) the location of the person and various objects, (b) whether those objects will be interacted with in the future, and (c) how will the person interact with each object. Besides assistive robotics~\cite{assistive,enrichme}, predicting future interactions could enable path planning and navigation~\cite{poni,objectnav,nav-ziad,nav-2}, imitation learning~\cite{human-to-humanoid,humanoid-wang,ilya-dexterous}, AI coaching and AR assistants~\cite{expertaf,shapiro-lights}, and view planning~\cite{scenediffusion,jiang2023symphonize3dsemanticscene}.

Current methods attempt to solve object interaction anticipation as a 2D video problem---inferring 2D heatmaps on frames, naming the next action, or detecting the likely next active object~\cite{hotspot-tushar,oct_hoi,interaction-2,interaction-3,interaction-4,afftention,avt,hiervl,ego4d-anticipation-3}. %
However, %
this common 2D formulation overlooks the persistent 3D context of the environment and object layout, instead viewing the world as detached glimpses from %
the camera's point of view.  %
Other work anticipates future 3D human poses %
while taking context from the scene~\cite{scene-context-1,scene-context-2,scene-context-3}, typically using autoregressive models~ \cite{4dhumans,t2m-gpt,chatpose}.  However, these methods are not 
interaction-centric, i.e., they focus on motion generation without attempting to model %
object interactions. 
Both lines of work %
typically predict only a few seconds into the future, within the same action (e.g., walking, chopping). %

Our key insight is that a person's movement and interaction are tightly linked to the activity they are doing and the objects in 
their environment. Therefore, it is crucial to address this task with %
3D knowledge about the persistent surrounding environment, its objects, and the person's body poses.
The activity and the objects in the surroundings impact both \emph{where} will be the future interactions and \emph{how} the person will interact in the environment. For example, if a person is making tea, %
it is likely that the person will interact with some water source to get water. Moreover, if the water dispenser is wall-mounted, the person will extend their arm to reach it. See Fig. \ref{fig:teaser}. %

To overcome the gap in today's models, we introduce a 4D formulation of the interaction anticipation problem, and we propose \modelname, a novel model to address it.  %
As input, our model takes a video observation of a person doing an activity, %
together with a 3D scene representation containing the  objects and the person.\footnote{The 3D scene representation can either be derived directly from the video using state-of-the-art methods~\cite{slam,orb-slam,droid-slam,deep-vis-slam}, or estimated from richer sensors when  available~\cite{aria-glass}.}  The proposed model encodes the past observations into a multimodal representation with a transformer, then learns to decode back to all future 3D interaction locations and, for each location, a distribution of likely body poses that will be executed there.
We hypothesize, and experimentally validate, that video and an explicit 3D scene context helps predict the interaction location (\emph{where}) and the human body pose at the time of interaction (\emph{how}).   Furthermore, considering interactions in 4D facilitates longer-horizon anticipation---imagining how the activity in a given physical space will unfold over the course of minutes, not just seconds, as a person moves about to different regions of the 3D environment.

We leverage the recently introduced Ego-Exo4D \cite{egoexo4d} dataset to create the training and the testing dataset for future interaction location prediction (\emph{where}) and future interaction body pose prediction (\emph{how}), thus capturing both essential aspects. 
In extensive experiments, we show that our method outperforms various related state-of-the-art methods in interaction anticipation---autoregressive action/pose/hotspot prediction \cite{hiervl,oct_hoi,4dhumans,t2m-gpt}, and video-based methods without explicit 3D environment conditioning \cite{voxformer,occformer}. Specifically, our method is more than 30\% better than %
the best performing baseline. %
Overall, this work offers an important step towards realizing the synergy between 3D environments and human action, which are often treated independently in the  literature to date.

%% file: sec/2_related_work.tex
\section{Related Work}
\label{sec:related-work}

\textbf{Learning about activity in videos.} 
Video datasets like HowTo100M \cite{howto100m}, Kinetics \cite{kinetics}, Ego4D \cite{ego4d} and Ego-Exo4D \cite{egoexo4d} contain videos of people doing a wide variety of activities %
and support various computer vision tasks including activity recognition \cite{omnivore,mvitv2,uniformer,memvit,slowfast}, procedural planning \cite{procedure2,procedure3,procedure-learning-fei-fei-li}, task graph learning \cite{task_graph,video-distant,task-graph-objective,graph2vid}, action anticipation \cite{avt,rulstm,intention,whenwillyoudowhat,gao2017red,ego4d-anticipation-1,ego4d-anticipation-2,ego4d-anticipation-3,ego4d-anticipation-4,ego4d-anticipation-5},
goal forecasting \cite{darko}, 
and representation learning \cite{hiervl,mil-nce,videoclip,egovlp}. %
Despite many promising results, there is a heavy emphasis on 2D video devoid of the underlying persistent 3D context.  At best, models are expected
to implicitly learn the spatial arrangement of objects in 3D. This fundamentally limits the spatial understanding of these models, particularly for long-horizon tasks like future interaction prediction.
In this work, we propose to use explicit 3D environment context as an essential input 
alongside the video frames.  We show this \emph{early-fusion} enables the model to understand the spatial nature of the activity, %
resulting in a superior performance over video-based models without 3D context or whose 3D context is infused post-hoc.

\textbf{Predicting object interactions.} Humans interact with objects in various ways, e.g., lifting, dragging, sitting, sleeping upon. Human object interaction \cite{lemon,cg-hoi,3d-aff,where2act,adaafford} focuses on understanding 
affordances and contact points for a given object. Likewise, hand-object interaction \cite{hoi-affordance,contactopt,hoi-xiaolong,graspit,grasp-pc} predicts the hand pose when performing actions like grasping or holding for a given object. However, all the prior work assumes a known object (out of environment context) that is the target interaction. Moreover, the interaction is instantaneous, typically spanning a few seconds.  In contrast, our setting entails 3D scenes with multiple objects, and we predict both things---the location of the objects that will be interacted with, and the corresponding body poses over the course of multiple minutes.

Hotspot anticipation \cite{hotspot-tushar,interaction-2,interaction-3,interaction-4,afftention} is a related problem of localizing interaction points as heatmaps in 2D video frames. %
However, this prior work %
has no persistent 3D spatial understanding. In contrast, we leverage 3D spatial information, extractable from the same egocentric video, %
to predict interactions registered in the underlying 3D space containing multiple objects over a long duration.
Unlike prior work that infers the affordance map of a static 3D environment~\cite{multi-label-affordances,interaction-landscape}, our work anticipates the actions a person will perform next given a partial video of their activity.

\textbf{Human pose from videos.} Extracting human body pose from images or video is an active area of research \cite{4dhumans,wham,tokenhmr,slahmr,smplx,smpl}. Typical output formats are SMPL \cite{smpl} or body skeletons \cite{coco}. %
Some work attempts to predict human poses in the future---using motion priors \cite{pose-prediction-1,pose-prediction-2,pose-prediction-3,pose-prediction-4,pose-prediction-5,humor} and 3D scene context \cite{scene-context-1,scene-context-2,scene-context-5,scene-context-6,scene-context-7,scene-context-9,yan2024forecasting} as additional information.  
These methods can generate plausible motion patterns for short-term actions---especially periodic ones like walking, running, or jumping---but they do not tackle anticipation of longer-term behaviors, namely actions extending beyond the current one that entail interacting with different future objects in different future steps of the activity.

Recent work \cite{scene-context-2,scene-context-3,scene-context-4,scene-context-5,scene-context-9} shows how humans would interact with an object in the scene, e.g., chair, sofa, staircase.  %
The datasets used for this task are mostly simulation-based \cite{scene-context-1,habitat3} or small-scale  \cite{prox-dataset,scene-context-3}.  %
Most importantly, these methods generate pose sequences for a fixed (singleton) object, whereas ours predicts future poses conditioned on past activity in a dynamic scene and in the presence of multiple potential interacting objects.  Our idea can be seen as ``interaction-centric pose anticipation", a new and important dimension of this problem space.

%% file: sec/3_method.tex
\section{Approach}
\label{sec:method}

\begin{figure*}[t]
    \centering
    \includegraphics[width=\linewidth]{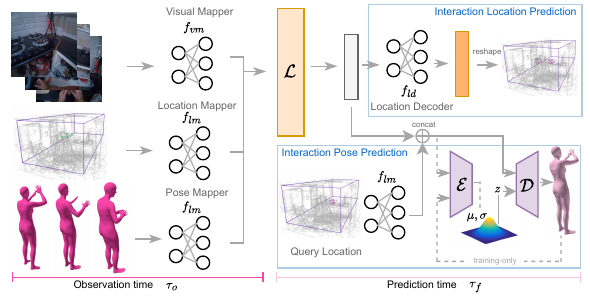}
    \caption{\textbf{Overview of the \modelname~approach.} The past observation information (video, pose, and environment) is encoded into a multimodal representation (left). The multimodal encoder $\mathcal{L}$ encodes the past observation, and is used to predict the interaction locations using a decoder (top right). We use the past observation encoding, along with the query location, to train a CVAE encoder decoder to generate a location-specific pose distribution conditioned on the past activity (bottom right). 
    }
    \label{fig:method} 
    \vspace{-0.1cm}
\end{figure*}

We first formally describe the problem (Sec. \ref{sec:problem}). Next, we describe the model design for our task (Sec. \ref{sec:model}), the dataset preparation strategy (Sec. \ref{sec:data-preprocess}), and implementation details (Sec.~\ref{sec:implementation}).

\subsection{Future interaction prediction}
\label{sec:problem}

Given an observation of a person performing an activity, up to a time instant, we want to anticipate \emph{all} subsequent interactions, up to a future time (3 minutes in our experiments). We capture both aspects of an interaction---its location (\emph{where}), and its body poses %
(\emph{how}).
We define interaction as making contact with an object of interest to use it, e.g., \emph{lifting a cup}, \emph{opening a fridge}, as opposed to non-contact actions, e.g., \emph{monitoring the oven}, \emph{observing the bike tire}.

\textbf{Future interaction location prediction:} Firstly, given an observation video $\mathcal{V}$ till time $\tau_o$, and the 3D locations, encoded as %
$\mathcal{P}$ (obtained from $\mathcal{V}$ or otherwise), 
we aim to learn a future object interaction function $\mathcal{F}_o$ that predicts all future object interactions, till time $\tau_f$. Following \cite{avt}, we allow for a small anticipation time $\tau_a$ buffer immediately following $\tau_o$. The output should be a set of points $\mathbf{x} \in \mathbb{R}^3$ such that the person interacts with the object at the 3D point $\mathbf{x}_{\tau_k}$ at the timestamp $\tau_k$, for all such interaction timepoints before $\tau_f$: %
\begin{align}
\label{eq:object}
    \mathcal{F}_o \left( \mathcal{V}_{0:\tau_o}, \mathcal{P}  \right) = \left\{ \mathbf{x}_{\tau_k}~|~ \mathcal{I}(\mathbf{x}_{\tau_k}) \in \mathcal{O} \right\},
\end{align}
where $\tau_f  > \tau_k > \tau_o + \tau_a$, and the interaction function $\mathcal{I}(\mathbf{x}_{\tau_k}) = o \in \mathcal{O}$ if the person is interacting with an object $o$ at location $\mathbf{x}$ at time $\tau_{k}$; $\phi$ otherwise. Here, $\mathcal{O}$ is the set of all objects.

\textbf{Future interaction pose prediction:} Next, at each interaction point, we want to predict the body pose. However, there are various correct possible future body poses for a given interaction point and an object. For example, a person can reach out to the faucet to \emph{turn it on}, or to \emph{clean it}, both having different body poses. Thus, we learn a  function to return the distribution of the likely body poses, rather than a deterministic body pose: 
\begin{align}
    \mathcal{F}_p \left( \mathcal{V}_{0:\tau_o}, \mathcal{P}, \mathbf{x}_{\tau_k} \right) = \mathbb{P}\left( \theta, t \right),
\end{align}
where we learn a distribution over the SMPL \cite{smpl} body pose parameters $\theta$ and the location $t$, for any given query location $\mathbf{x}_{\tau_k}$. %
We omit the SMPL shape parameter $\beta$, since it does not change during interactions.
Likely %
body poses are obtained by sampling from %
$\mathcal{F}_p$.

\subsection{Learning future interactions}
\label{sec:model}

Next we introduce the training framework for learning the future object interaction $\mathcal{F}_o$ and future pose distribution $\mathcal{F}_p$ functions. The common input to both these functions is the video observation till time $\tau_o$. We use the following %
components for observation input: the egocentric video of the actor $V$, the body pose of the actor $P$, and the object and the actor placement in the 3D scene. %
We obtain a common modality-agnostic representation for all the inputs and feed it into a multimodal transformer. The transformer encoder output is then used to predict the future interaction locations. For future pose distribution prediction, we additionally provide a location as a query %
to a CVAE \cite{cvae} based encoder-decoder. The architecture is shown in Fig.~\ref{fig:method} and detailed below.

\textbf{Representing the 3D scene as voxels.} We discretize the 3D space as a $N \times N \times N$ voxel grid. The boundaries are chosen as the maximum of the actor movement and the object placements. %
We choose a discretized, explicit voxel representation for ease of training and interpretability, as opposed to alternatives based on implicit NeRF~\cite{nerf}-like functions.

\textbf{Encoding egocentric actor observation.} Given an egocentric %
observation till time $\tau_o$, we use an off-the-shelf feature extractor $f_V$ to obtain video representation $f_V(V)$. The visual feature extractor is a large pretrained encoder network that offers good semantics about the activity. We use a visual mapper $f_{vm}$ to output a modality-agnostic representation $\mathbf{\bar{v}} = f_{vm}(f_V(V))$ that can be fed to a transformer. The visual mapper \cite{llava,llava-med} is used to convert visual features to a representation that can be directly used with a transformer model.

\textbf{Encoding body pose observation.} Next, we encode the observed SMPL \cite{smpl} body pose parameter $\theta$ of the pose $P$. 
These are real-valued numbers, and we use a pose mapper $f_{pm}$ to convert the pose parameters into a representation $\mathbf{\bar{p}}~=~f_{pm}(\theta)$. The location $t$ is encoded along with the object bounding boxes, explained next, for uniformity.

\textbf{Encoding object bounding boxes.} We represent the 3D scene $S$ as voxels, as described above. We assign an index $i$ at a voxel location containing object $o$ if $o = \mathcal{O}[i]$, where $\mathcal{O}$ is the object taxonomy \cite{lvis}. We use a reserved index to denote the actor location $t$ in this scene $S$. Next, we use a location mapper $f_{lm}$ to encode the object representations and actor locations as $\mathbf{\bar{o}}~=~f_{lm}(S)$, same as above.

\textbf{Multimodal transformer encoder.} Next, we take the multimodal representations $\mathbf{\bar{v}}, \mathbf{\bar{p}}$ and $\mathbf{\bar{o}}$ and use a transformer with self-attention to learn an output representation, $\mathbf{\bar{r}} = \mathcal{L}(\mathbf{\bar{v}}~|~\mathbf{\bar{p}}~|~\mathbf{\bar{o}})$. We choose the first output of the transformer as the output representation, following standard practice \cite{timesformer,taco}. The output representation $\mathbf{\bar{r}}$ encapsulates the observation information till time $\tau_o$.

\textbf{Decoding future interaction location.} The final stage of learning $\mathcal{F}_o$ involves finding \emph{all} the future interaction locations. We use the same $N \times N \times N$ voxel representation at the output. A voxel is marked $1$ when the corresponding location has a future interaction; $0$ otherwise. (We detail how to extract these voxel maps from unlabeled video in Sec.~\ref{sec:data-preprocess}.)  To achieve this, we use a simple location decoder linear layer $f_{ld}$ that maps $\mathbf{\bar{r}}$ to a vector having $N^3$ dimensions. The output is then reshaped to $N \times N \times N$ grid to obtain the predicted future interaction location grid $\hat{L}$.

\textbf{CVAE for future pose distribution.} In addition to the output representation $\mathbf{\bar{r}}$, the pose distribution function $\mathcal{F}_p$ requires an additional input for any query location $x \in \mathbb{R}^3$ that denotes the location at which the pose distribution is desired.
We use the same voxel representation, and the location mapper $f_{lm}$ to encode the reference location as $\mathbf{\bar{x}}$.

Following the standard CVAE \cite{cvae} architecture, the CVAE encoder $\mathcal{E}$ takes as input the multimodal output representation $\mathbf{\bar{r}}$, the location embedding $\mathbf{\bar{x}}$, and the desired pose output $P \sim \mathbb{P}(\theta, t)$ and outputs the latent distribution parameters $\mu, \sigma$, i.e. $\mu, \sigma = \mathcal{E}(\mathbf{\bar{r}}, \mathbf{\bar{x}}, P)$. Here, $P$ is a sample from the pose distribution. Next, the decoder $\mathcal{D}$ tries to reconstruct the sampled pose $P$ using the inputs $\mathbf{\bar{r}}, \mathbf{\bar{x}}$ and a sample $z$ such that $z \sim \mathcal{N}(\mu, \sigma)$, i.e. $\hat{P} = \mathcal{D}(z, \mathbf{\bar{r}}, \mathbf{\bar{x}})$. At inference, we can sample multiple $z \sim \mathcal{N}(0, 1)$ and use that to predict output pose samples, $\hat{P} = \mathcal{D}(z, \mathbf{\bar{r}}, \mathbf{\bar{x}})$. We use $\beta$ from the past observation, alongside the predicted $\theta, t$, to reconstruct the SMPL body mesh/3D joints.

\textbf{Training objectives.} 
We now describe the training objective to learn the functions $\mathcal{F}_o$ and $\mathcal{F}_p$.
As described in Sec. \ref{sec:model}, we represent the future location as a binary voxel: $1$ if it contains a future interaction, and $0$ otherwise. Consequently, we learn $\hat{L}$ as the output of the function $\mathcal{F}_o$. Similarly, we represent the ground truth output locations $\{\mathbf{x}_{\tau_k}\}$ (Eq. \ref{eq:object}) as a voxel grid $L$.
Thus, we use the standard binary cross entropy (BCE) loss for the future interaction location prediction task.

Next, for the future pose distribution prediction task, we use a combination of reconstruction loss and the KL divergence between the predicted distribution parameter and the standard normal. For the reconstruction loss, we use the MSE loss between the predicted SMPL parameters $\hat{P}~=~(\hat{\theta}, \beta, \hat{t})$ and the ground truth parameters $P~=~(\theta, \beta, t)$. Moreover, following \cite{4dhumans}, we also convert the SMPL parameters to 3D body joints, $J = \text{SMPL}(P)$, and compute the joint $L_1$ error. Finally, the KL divergence error is computed between the predicted parameters $(\mu, \sigma)$ and $(0, 1)$. Overall, the training objective is to minimize
\begin{align*}
    w_{S} \lVert P - \hat{P}\rVert_2 + w_J \lVert J - \hat{J}\rVert_1 + KL\left(\mathcal{N}(\mu, \sigma), \mathcal{N}(0, 1)\right),
\end{align*}
where the weights $w_S$ and $w_J$ control the loss contributions. %

\subsection{Future interaction dataset}
\label{sec:data-preprocess}

As introduced %
above, we use the egocentric video $V$ and body pose $P$ as the modalities in this task, and we also use the object interaction locations in the 3D scene. %
While no existing dataset provides exactly this prepared data, 
Ego-Exo4D \cite{egoexo4d} offers the necessary raw data and serves as an excellent large-scale, diverse testbed for our work (see Sec.~\ref{sec:implementation} for scope and statistics).  
Ego-Exo4D data is recorded with Aria glasses~\cite{aria-glass} having dedicated SLAM cameras, thus enabling superior 3D registration, %
an advantage over %
using RGB frames for 3D information extraction \cite{sfm,vins}, as done in \cite{epic-aff,epic-fields}. 
We find the object placements and human poses %
as follows.

\textbf{3D object bounding boxes.} Object segmentation in 3D \cite{segcloud,interactive-seg-3d} and 2D \cite{detic,sam} is an active area of research.
Ego-Exo4D's SLAM cameras provide a mapping between the video pixels to the 3D locations. Therefore, we perform object segmentation in 2D video frames using Detic \cite{detic} (that uses the object taxonomy from LVIS \cite{lvis}) and use the above mapping to convert segmented pixels to object point clouds. %
Next, we use DBSCAN \cite{dbscan} for density-based clustering to find the count of an object %
and make the bounding boxes tight. Lastly, we find an oriented bounding box \cite{obb} for every object in the scene. See Supp.~for more implementation details and visualizations.

\begin{figure}
    \centering
    \includegraphics[width=0.9\linewidth]{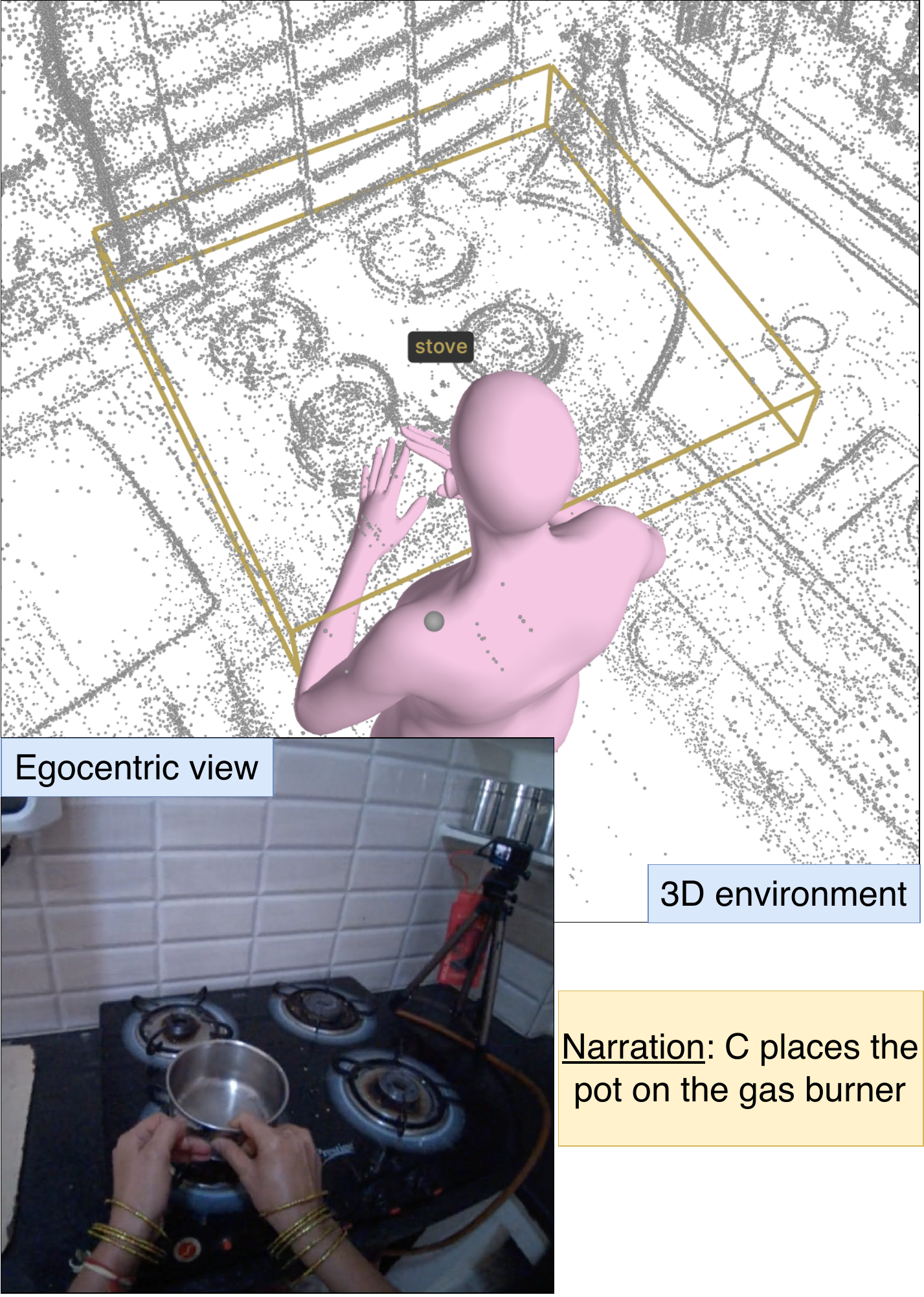}
    \caption{\textbf{An interaction instance.} We mark a timestamp as an interaction when the hands are within the 3D bounding box of an object referenced in the narration. An LLM is used to match the narration with the objects in the detector's vocabulary, e.g., \emph{stove} and \emph{gas burner}.}
    \label{fig:interaction-example}
    \vspace{-0.3cm}
\end{figure}

\textbf{Extracting body pose.} %
We use the state-of-the-art WHAM~\cite{wham} model to obtain body pose and shape from videos. The method first tracks the person in the video \cite{vitpose}, followed by a regression head that predicts the SMPL parameters ($\theta, \beta, t$). Even though the Ego-Exo4D dataset assumes one actor per video, there are additional people present in the video that are tracked by WHAM; see Supp. for examples, and how we disambiguate the actor. Also, though the dataset contains multi-view videos, we use only the view having the maximum joint visibility for extracting SMPL. %
Lastly, we extract all the body poses in the local coordinate system, and use the egocentric %
parameters to place the person in the 3D global coordinate frame. %

\textbf{Finding interaction instances.} After obtaining the body pose and object bounding boxes, the final stage of the data preparation involves finding the interaction timestamps. One potential approach is to simply find the intersection points of the body pose hands with the object bounding boxes. However, %
this would be susceptible to minor errors in the estimated 3D object bounding boxes and body pose parameters.  Hence, for a good quality dataset, we propose a geometric video-language approach.  We take the natural language ``narrations" describing each action in Ego-Exo4D and their accompanying timestamps.  
Then we use Llama-3.1-8B \cite{llama-herd} to classify all the narrations into either a touch-based interaction or a non-touch interaction, and to match the object mentioned in the narration to the object detection vocabulary \cite{lvis}---a task well-suited for a large language model. For example, \emph{``the person picks up a metal skillet holder''} shows an interaction with a skillet, whereas \emph{``the person watches the skillet"} does not. See Supp.~for the prompt details. Using these outputs, a final interaction instance is when either of the person's hands is within the 3D object bounding box \emph{and} the LLM marks a timestamp as a touch-based interaction. See Fig. \ref{fig:interaction-example}. %

 In summary, we use videos and state-of-the-art methods to extract objects, body poses, and interaction instances, all registered in the same coordinate space.  While our implementation takes advantage of the high quality camera calibration offered in Ego-Exo4D, as visual SLAM continues to improve~\cite{slam,orb-slam,droid-slam,deep-vis-slam}, such an approach will generalize increasingly better to lower quality data as well.  
 We %
emphasize that ours is the first work to curate a dataset for future interaction prediction in 4D videos.  We are sharing our data to facilitate benchmarking by other researchers.

\subsection{Implementation details}
\label{sec:implementation}

\begin{figure*}
    \centering
\includegraphics[width=\linewidth]{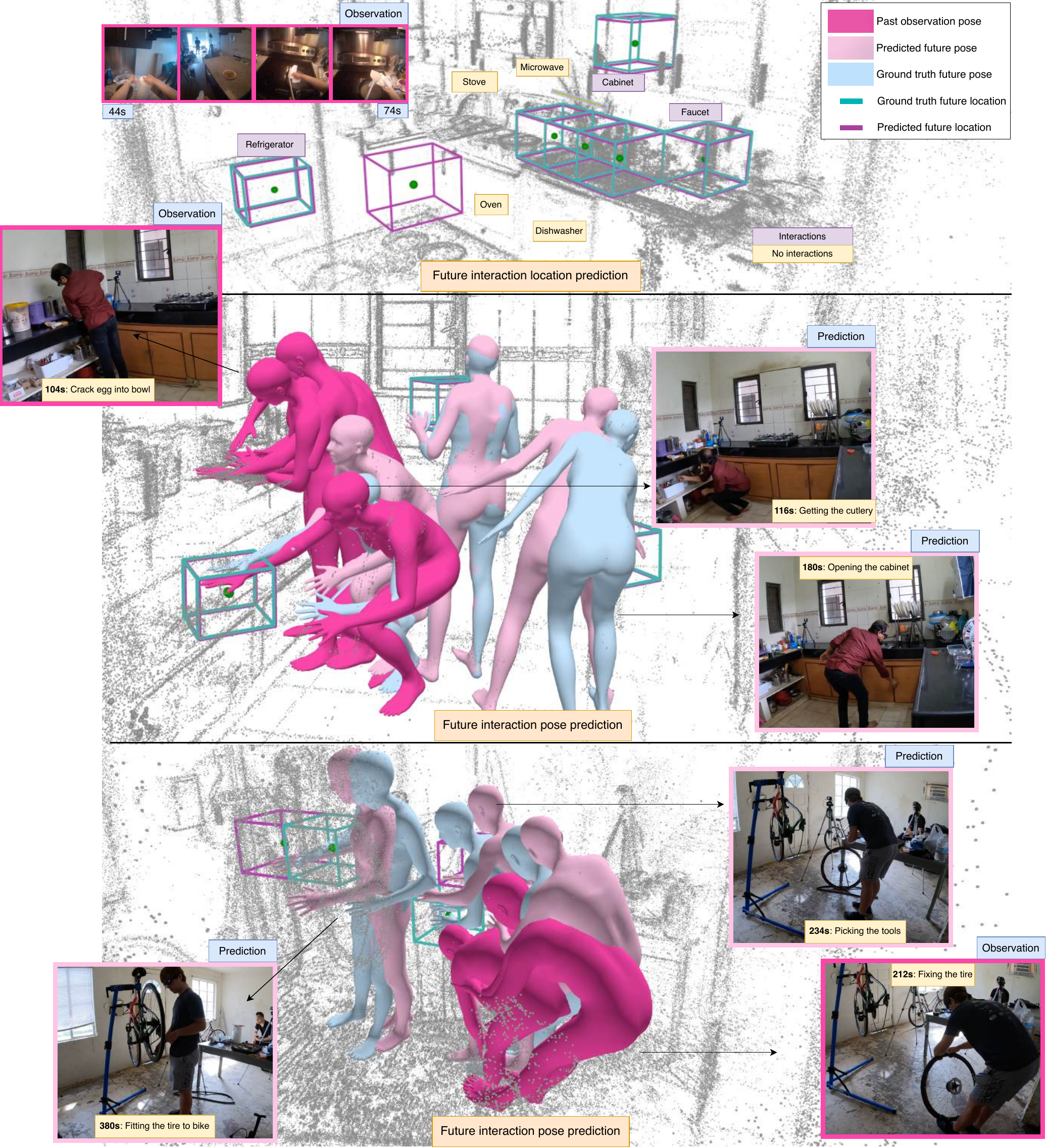}
    \caption{\textbf{Visualization of example results.} (Top) Interaction location prediction in a cooking take. Based on the observed input ego, the model is able to correctly predict future interaction locations--- \emph{refrigerator, faucet, cabinet}. (Middle, bottom): Interaction sequences in a cooking and a bike repair take. Given the past observation information, our model is able to accurately predict future interactions. If the observed sequence shows the person fixing the tire, the model predicts that the tire will be fixed onto the bike at a later stage, at a different spatial location (bottom, left most pose visualization). Best viewed in zoom. Exo view shown for visualization only. %
    }
    \label{fig:main-result}
\end{figure*}

\textbf{Dataset and statistics.} Ego-Exo4D \cite{egoexo4d} contains 5,035 takes covering physical and procedural scenarios. Physical scenarios like soccer and basketball are typically performed outdoors and/or contain limited object interaction. Thus, we focus on the procedural scenarios---\emph{cooking}, \emph{bike repair}, and \emph{health}, which are comprised of 196,363 total instances of %
interactions (e.g., \emph{pick up cup}, \emph{tighten chain}, \emph{twist open bottle}, etc.), and span 92 unique environments.  Each take has, on average, 130 interactions, with 18 distinct object interactions.  
The average length of a take is 5.75 minutes. Following our task definition (see Sec. \ref{sec:problem}), 
we set $\tau_o = 30s, \tau_a = 5s, \tau_f = 180s$. %

This dataset creation strategy results in 87,373 training (cooking: 62,495, bike repair: 13,589, health: 11,289), 10,563 validation (cooking: 7,837, bike repair: 1,478, health: 1,248), and 10,124 test episodes (cooking: 7,197, bike repair: 771, health: 2,156).
We apply the same take-level train/test split as in \cite{egoexo4d}. 
The environment encountered at test time is never observed during training.
Across cooking, bike repair, and health, each episode has an average of 10, 20 and 6 future interactions and 3.5 different body poses per interaction location, respectively. %

\textbf{Network architecture and parameters.} $f_V$ is the EgoVLPv2 \cite{egovlpv2} visual encoder. We use pre-extracted features for a faster I/O. All the mappers $f_{vm}, f_{pm},$ and $f_{lm}$ are linear layers, with the same output dimension, i.e., 4096. The input dimension to $f_{vm}$ is 4096, consistent with the output of $f_V$. Similarly, $f_{pm}$ has 207 ($23\times 3 \times 3$) input dimension representing the SMPL pose $\theta$,  and $f_{lm}$ represents a location in the $N\times N\times N$ grid, i.e., $3N$ input dimension. We use $N=16$ to convert the 3D space into voxels. This choice reasonably captures the objects in the environment, while being experimentally feasible (i.e., fits the GPU memory). The method can be directly applied
to any $N$, with the model parameters scaling as $N^3$, same as other voxel-based methods \cite{voxformer,occformer}.  All modalities are processed at $1$ feature per second. We train the distinct scenarios separately (cooking, health, bike repair). We train all the models on eight Quadro RTX 6000 23GB GPUs. We set a learning rate of $5\times10^{-5}$ for $3$ epochs for predicting interaction location and $5 \times 10^{-6}$ for $6$ epochs for predicting interaction pose. All runs take a maximum of $10$ hours.

%% file: sec/4_experiments.tex
\section{Experiments}
\label{sec:expts}

\input{tables/result_table}

We first describe the baselines, %
ablations, and %
metrics. %
Then, we discuss our experimental results and comparisons, along with some output visualizations.%

\textbf{Baselines.} We compare %
to a total of 6 state-of-the-art models developed for related tasks. %
While these models are all relevant, none of them exactly performs 4D future interaction prediction, so we appropriately modify the existing models to create strong baselines organized into two families: autoregressive and video-to-3D.
The family of autoregressive baselines (\textbf{HierVL}~\cite{hiervl} for long-term anticipation, \textbf{OCT}~\cite{oct_hoi} for hotspot prediction, \textbf{4D-Humans}~\cite{4dhumans} and \textbf{T2M-GPT}~\cite{t2m-gpt} for pose generation/prediction) are transformer-based models that use the observed context to generate the next token for a given modality (e.g., action label, next frame heatmap, or future body pose).  
For fair comparison to our model, we use the camera parameters in Ego-Exo4D to lift the 2D models' predictions of interaction points into the 3D environment.
The video-to-3D models (\textbf{VoxFormer}~\cite{voxformer}, \textbf{OccFormer}~\cite{occformer}, \textbf{Video-to-pose} CVAE~\cite{cvae}) directly predict the 3D location or pose as output from video, hence implicitly learning the 3D semantics of the scene.  VoxFormer~\cite{voxformer} and OccFormer~\cite{occformer} are originally trained to predict 3D scene occupancy from images, so we adapt to video.  We train all the video-to-3D prediction models on our data and explore fine-tuning T2M-GPT-FT~\cite{t2m-gpt} on our dataset as well (\textbf{T2M-GPT-FT}).  Please see the Supp.~for all implementation details.

\textbf{Ablations.} In addition to the baselines, we also compare against various ablations of the architecture design, where we remove each component of our method's input in turn to discern their impact: the observed egocentric video (\textbf{\modelname~w/o video}), the person's body poses (\textbf{\modelname~w/o pose}), and the environment context, i.e., object layout (\textbf{\modelname-w/o env}).  See Supp.~for ablations with hyperparameters.

\textbf{Metrics.} For future interaction location prediction, we use Chamfer distance and PR-AUC (precision-recall area under curve).  
Chamfer distance is lower for a better method and reported in voxel units, whereas PR-AUC ranges in $[0, 1]$, higher the better (shown out of $100$ for clarity).  For future interaction pose prediction, we report Mean Per Joint Position Error (MPJPE) in world-coordinates, and  Procrustes Aligned MPJPE (PA-MPJPE), following prior work \cite{4dhumans}. Both metrics measure the per-joint error in mm, lower the better. Recall that we predict a distribution of the pose parameters for a given object interaction location. At inference, we sample $N=5$ poses, and report the error of the closest pose w.r.t. the ground truth, same for any baseline that outputs multiple poses such as T2M-GPT.  %

\textbf{Results.} Tab. \ref{tab:results} (left) shows the results for future interaction location prediction. Our method significantly outperforms all prior work on all scenarios for both the PR-AUC and Chamfer metrics, with relative gains more than 32\% (absolute 4.6\%). In particular, autoregressive models (HierVL \cite{hiervl}, OCT \cite{oct_hoi}) are good at predicting the next few interaction locations, but the errors in the prediction accumulates and the location often diverges. Similarly, video-to-3D future prediction models (Occformer \cite{occformer}, VoxFormer \cite{voxformer}) also fail to learn the future location due to their lack of an explicit activity and 3D reasoning. 

Comparing with ablations, providing both the video and pose sequence improves the performance over a single modality. The performance drops significantly if the location context of the environment is not provided---showcasing its importance. %
Note that other autoregressive methods are given location information as \emph{late-fusion}. The performance in health (COVID-19 test) is lower than other scenarios due to fewer interactions with objects in the chosen off-the-shelf object vocabulary \cite{lvis}, meaning limited context is passed to the model. 

Tab. \ref{tab:results} (right) shows the results for future interaction pose prediction. Our method outperforms all baselines, by up to 49 mm. Same as above, the autoregressive methods (4D-humans~\cite{4dhumans} and T2M-GPT~\cite{t2m-gpt}) diverge when required to generate very long (3 min) pose sequences. Finetuning T2M-GPT marginally improves the performance, with the autoregressive nature of sequential pose generation being the limiting factor. Similarly, the vanilla video-to-pose CVAE is unable to capture the future poses, due to the missing environment context.  We observe the same trend as above when comparing with the ablations. Note that in case of `w/o pose', the model is still strong due to the other modalities---environment context, including the actor's location, and the video observation.

Finally, Fig.~\ref{fig:main-result} shows some qualitative results. Our method is able to predict the correct interaction locations across different parts of the environment. Given the past observation of cleaning a pan, the model predicts fetching veggies from the refrigerator and cabinet, and using the faucet (top). The model also predicts that microwave, stove, among other things, will not be interacted with in the next 3 minutes. We also see how our method can identify the current activity and predict the future pose, e.g., if the person is fixing a bike tire, the person will likely interact with the bike to put the tires back on (see Fig. \ref{fig:main-result}, middle and bottom). For a qualitative comparison between our method and baselines, please see Supp. Fig.~\ref{fig:baseline}.

Overall, these strong results suggest the effectiveness of our proposed model for future interaction prediction, and our dataset and evaluation paradigm establish a valuable benchmark for continued work.

%% file: tables/result_table.tex
\begin{table*}[ht]\footnotesize
\begin{minipage}{0.5\textwidth}
    \centering
    \begin{tabular}{L{2.0cm}C{0.5cm}C{0.6cm}C{0.5cm}C{0.6cm}C{0.5cm}C{0.6cm}}
        \multicolumn{7}{c}{\textbf{Future interaction location prediction}} \\ \hline
         & \multicolumn{2}{c}{\textbf{Cooking}} & \multicolumn{2}{c}{\textbf{Bike Repair}} & \multicolumn{2}{c}{\textbf{Health}} \\ 
        \cmidrule(lr){2-3} \cmidrule(lr){4-5} \cmidrule(lr){6-7}
        & PR & Ch $\downarrow$ & PR & Ch $\downarrow$ & PR & Ch $\downarrow$ \\ \hline
        HierVL \cite{hiervl} & 11.2  & 11.0  & 10.6 & 11.3  & 7.8 & 51.3  \\ 
        OCT \cite{oct_hoi} & 16.9  &  9.0 & 13.3 & 9.4 &  11.2& 44.8 \\
        OccFormer \cite{occformer} & 13.6  & 9.8  & 14.0 &   10.0 & 10.3 & 46.5 \\
        VoxFormer \cite{voxformer} & 15.1  &  9.5 & 14.1 & 10.5  & 9.9 & 46.5 \\
        \rowcolor{Gray}
        \modelname & \textbf{21.0} & \textbf{7.4} & \textbf{18.7} & \textbf{7.2} & \textbf{12.7} & \textbf{41.7} \\
        ~~~w/o video & 20.0 & 7.6 &\textbf{18.7} & 9.2  & 12.0 & 43.7 \\
        ~~~w/o pose & 19.1 & 7.7 & 18.3 & 7.6 & 11.7 & 43.4 \\
        ~~~w/o env & 9.9 & 11.4 & 6.0 & 13.2 & 4.7 & 56.0 \\
        \hline
    \end{tabular}    
\end{minipage}
\begin{minipage}{0.5\textwidth}
\centering
        \begin{tabular}{L{2.4cm}C{0.5cm}C{0.6cm}C{0.5cm}C{0.6cm}C{0.5cm}C{0.6cm}}
        \multicolumn{7}{c}{\textbf{Future interaction pose prediction}} \\ \hline
         & \multicolumn{2}{c}{\textbf{Cooking}} & \multicolumn{2}{c}{\textbf{Bike Repair}} & \multicolumn{2}{c}{\textbf{Health}} \\ 
        \cmidrule(lr){2-3} \cmidrule(lr){4-5} \cmidrule(lr){6-7}
        & M $\downarrow$ & PA $\downarrow$ & M $\downarrow$ & PA $\downarrow$ & M $\downarrow$ & PA $\downarrow$  \\ \hline
        4D-Humans \cite{4dhumans} & 473 & 65  & 492  & 125 & 307 & 72  \\
        T2M-GPT \cite{t2m-gpt} & 397 & 65 & 470 & 117 & 276 & 70 \\
        T2M-GPT-FT \cite{t2m-gpt} & 264 & 61  & 410  & 99  & 221 & 66 \\
        Video-to-pose & 267 & 60 & 402 & 97 & 226 & 66 \\
        \rowcolor{Gray}
        \modelname & \textbf{229} & \textbf{56} & \textbf{372} & \textbf{91} & \textbf{172} & \textbf{62} \\ 
        ~~~w/o video & 234 & 58 & 375 & 92 & 175 & 68 \\
        ~~~w/o pose & 236 & 60 & 382 & 95 & 176 & 66 \\
        ~~~w/o env & 336 & 63 & 475 & 93 & 280 & 67 \\
        \hline
    \end{tabular}
\end{minipage}
    \caption{\textbf{Results of future interaction prediction.} We show the results for both future interaction location prediction (left) and future interaction pose prediction (right) for three scenarios---cooking, bike repair, and health. We outperform all prior work on both the tasks in all scenarios. See text for details. (PR: precision-recall area under curve, Ch: chamfer distance, M: MPJPE, PA: PA-MPJPE.)}
    \vspace{-0.5cm}
    \label{tab:results}
\end{table*}

%% file: sec/5_conclusion.tex
\vspace{-0.1cm}
\section{Conclusion}

We propose \modelname, a novel method to predict future interactions. We use video observations, along with the 3D scene context, to anticipate the location of the interaction and the person's body pose during that future interaction. We design a multimodal architecture to capture the activity intent and the person's location in the 3D scene. Our performance improvement over state-of-the-art work addressing related problems showcases the effectiveness of our approach and the novelty of the task itself. In the future, we plan to explore generalizations for predicting the movements of dynamic objects and a streaming variant that could repeatedly revise its future predictions.

%% file: sec/X_suppl.tex
\clearpage
\setcounter{page}{1}
\maketitlesupplementary

\appendix

\section{List of supplementary materials}

We attach a supplementary video containing an overview of the paper, including dataset and result visualization. We will also release the interaction dataset and code.

\section{Future interaction dataset}

This section contains additional details about the future interaction dataset, discussed in Sec. \ref{sec:data-preprocess}.

\textbf{3D object bounding boxes.} We use Detic \cite{detic}, along with the object taxonomy from LVIS \cite{lvis}, to find the mapping between the pixels in the video frames and the object labels. Since the inference from this method is fast, we perform segmentation at the original frame rate, i.e., 30 frames per second. Note that each pixel on the SLAM camera has an associated 3D location, which we use to map the object labels to a point in the 3D space. We perform object segmentation on SLAM frames directly because they have a direct mapping from 2D to 3D.

The DBSCAN \cite{dbscan} algorithm mentioned in Sec. \ref{sec:data-preprocess} is useful in tightening the bounding boxes. For example, if there are two chairs in the scene, attempting to create a bounding box directly results in the box containing everything between the two chairs. Thus, we use DBSCAN to find the approximate bounding boxes. We set $50cm$ as the threshold for distinct clusters, and require $100$ points at least to register as a unique object. This choice can correctly capture most of the objects seen in the chosen scenarios.

\textbf{Extracting body poses.} As mentioned in Sec. \ref{sec:data-preprocess}, the dataset has only one actor per video. However, there are other people present in some views. They are either bystanders or data collection volunteers. The dataset does not provide a full coverage of the annotation of the main actor in the videos. Thus, we use our heuristics to use the multi-view and disambiguate the main actor. Furthermore, the dataset contains multi-view videos showing the same person. We use the following two observations to extract the human pose. Firstly, the camera rig in the capture setup ensures the actor will have the largest area in all the video frames. Secondly, the similarity of the poses of the same person from all views will be higher than with people in the background. We use these observations to find the actor and then choose the \emph{best} view---having the maximum joint visibility---to obtain the extracted body pose. An alternative is to focus on maximum hand visibility. However, we do not over-emphaisze on the hands. Furthermore, comparing to manually annotated 3D poses in Ego-Exo4D (available for only a subset of the data), the MPJPE error is 82mm when we use max joints and 115mm when we use maximum hand visibility.

\textbf{Finding interaction instances.} We use the following prompt for Llama 3.1-8B \cite{llama-herd}:

\begin{tcolorbox}[breakable, boxrule=0.2mm]

\textbf{System:} You are a helpful AI assistant. Match the narrations with the object labels that is provided.

\textbf{User:} You are given narrations labeled by human annotators for a video. You are also given a set of object labels as per an object detection vocabulary. Find all instances of object interaction where the person would touch an object and map it to all the synonyms or similar words in the vocabulary. Sentences like `C looks at the fridge' has no object interaction. Objects like cup, glass can be grouped together. Here are the object labels that you have to use: \{labels\}. Answer in this format: 

1. \{rewrite first narration\} - 
answer: (object1, object2)

2. \{rewrite second narration\} - answer: NO INTERACTION

3. \{rewrite third narration\} - answer: NO MATCHING OBJECTS. Use `NO INTERACTION' and `NO MATCHING OBJECTS' in cases with no interaction and matching objects, respectively. Here are the numbered narrations: 

\{narrations\}
\end{tcolorbox}

\section{Details of baseline implementation}

We introduce the baselines in Sec. \ref{sec:expts}. None of the baselines are directly applicable for 4D interaction prediction. Thus, we appropriately modify related models to create strong baselines for comparison. The model and task-specific adaptations are listed below:

\begin{itemize}
    \item \textbf{HierVL \cite{hiervl}} is a recent method in Ego4D~\cite{ego4d} long-term anticipation (LTA) benchmark with publicly available codebase. This LTA version generates future action labels (nouns and verbs). We use the output noun and locate the same in the 3D space, and mark all voxels for the predicted object as future interaction locations. Since HierVL is initially pretrained on Ego4D, we do not need to finetune the dataset since the egocentric videos are from a similar distribution. We do, however, finetune the last layer to match the output class dimension to the objects detected in Ego-Exo4D scenes.
    \item \textbf{OCT \cite{oct_hoi}} is a recent work in joint hand motion and interaction hotspot prediction from EPIC-Kitchens-100 \cite{epic-kitchens-100}. We use this method to predict future interaction hotspot for the next 3 minutes. We then use the camera parameters in Ego-Exo4D to map the 2D interaction points into the 3D environment. Since, this model is also trained on egocentric videos, and just requires images as input, we do not retrain this method on Ego-Exo4D.
    \item \textbf{OccFormer \cite{occformer}} and \textbf{VoxFormer \cite{voxformer}} are methods originally designed for occupancy map prediction. We replace the image encoder in these networks with the video encoder $f_V$, used in our method.

    \item \textbf{4D-Humans \cite{4dhumans}} and \textbf{T2M-GPT \cite{t2m-gpt}} are recent works with autoregressive pose prediction capabilities. 4D-Humans extracts body pose from images and videos. We use the pose prediction module that predicts the next pose given the current body pose. We use this transformer autoregressively to generate multiple possible poses in the future. Similarly, T2M-GPT converts the body pose into a VQ-VAE based tokens and then predicts the pose tokens. The model is originally designed to generate pose based on the text condition; we modify the model to input prior pose tokens. Since our focus is not on \emph{when} a pose is happening but rather \emph{where}, we generously choose the prediction as the closest pose to the ground truth interaction location, out of all the generations. Both the methods are trained on large-scale pose datasets \cite{human36m,amass}. Regardless, we finetune T2M-GPT (called \textbf{T2M-GPT-FT}) on our dataset to investigate the role of the training data. We choose to finetune the latter model due to a better performance and the stable nature of VQ-VAE codebooks for pose token generation. 

    \item \textbf{Video-to-pose} CVAE \cite{cvae} model takes as input the video of the person and generates a future pose distribution. We use the same video encoder $f_V$ but do not provide any additional 3D context and expect the model to learn the 3D semantics implicitly. We train this method on our dataset. At inference, we choose the pose closest to the ground truth location.
\end{itemize}

\begin{figure}[t]
    \centering
    \includegraphics[width=\linewidth]{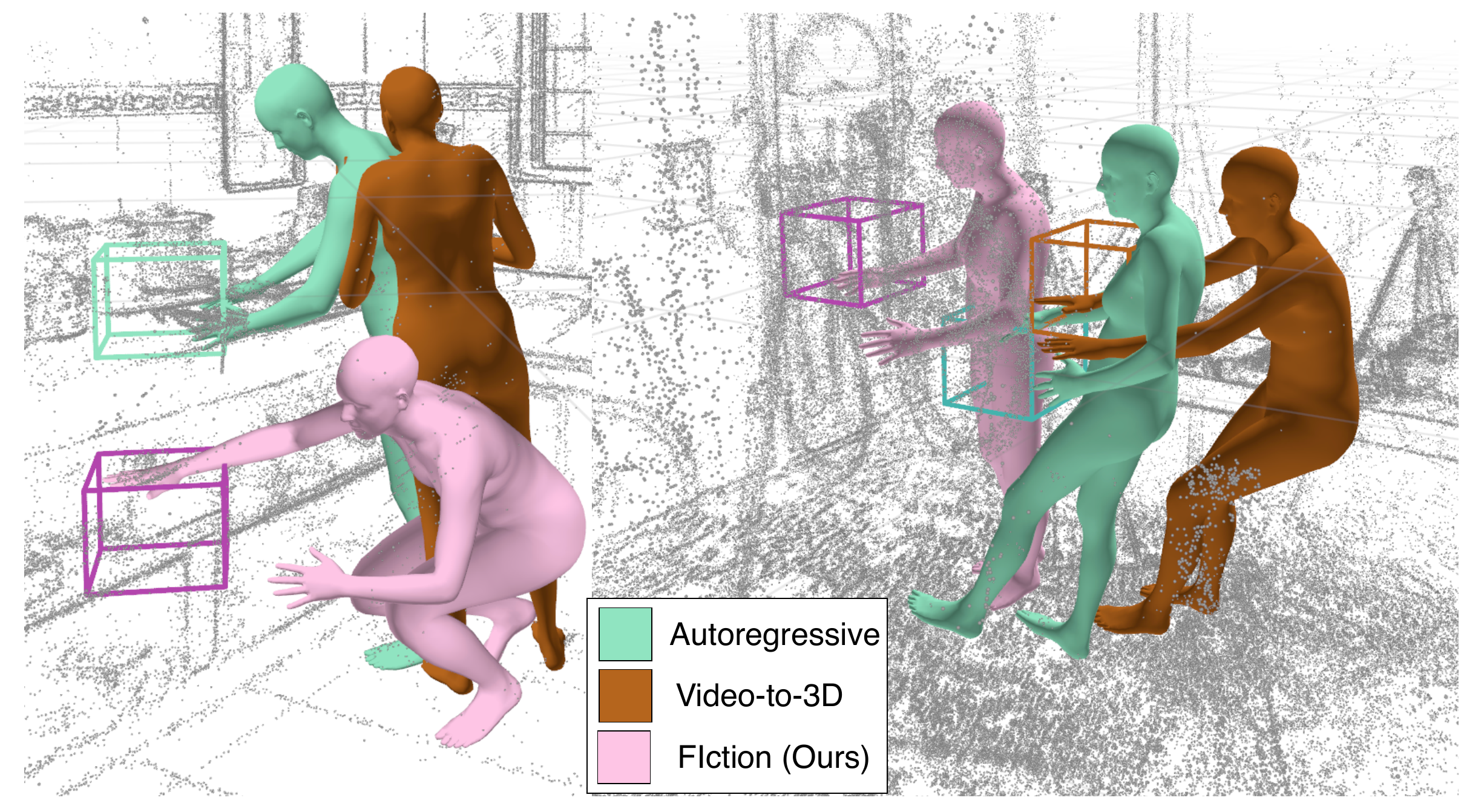}
    \caption{Comparison of our method with baselines and a cooking (left) and a bike-repair (right) take.}
    \label{fig:baseline}
\vspace{-0.5cm}
\end{figure}

\textbf{Qualitative comparison with baselines.} Fig. ~\ref{fig:baseline} compares our output with baselines. We see that our method is able to predict the interaction location and pose better than both the baselines. Autoregressive methods cannot predict long-term change in location and pose, while video-to-3D additionally misses the correct environment context.

\section{Additional ablations}

We also experiment with different choices of  hyperparameters. We only report numbers on training performed on the cooking scenario. The numbers are reported for the validation split, distinct from the testing split, mentioned in Sec.~\ref{sec:expts}. We only report PR-AUC and MPJPE for location prediction and pose prediction, respectively.

\textbf{Effect of the observation time $\tau_o$.}  Table~\ref{tab:ablation-observation-time} shows the results. We see that past video observation is crucial for providing the activity context. Thus, not providing any location gives the worst performance. The performance with $30$ seconds of past observation is at par with other past observation durations. Therefore, we choose $\tau_o=30$ so that the model has enough context for interaction prediction.

\begin{table}[h!]\footnotesize
    \centering
    \begin{tabular}{cc}
        Location prediction & Pose prediction \\ %
        \begin{tabular}{cccc}
             $0s$ & $30s$ & $60s$ & $120s$ \\ \hline
            16.0 & \textbf{21.2} & 21.0 & \textbf{21.2} \\ \hline
        \end{tabular}
        &
        \begin{tabular}{cccc}
             $0s$ & $30s$ & $60s$ & $120s$ \\ \hline
            225 & 215 & 213 & \textbf{212} \\ \hline
        \end{tabular}
    \end{tabular}
    \caption{Effect of $\tau_o$ on the performance.}
    \label{tab:ablation-observation-time}
\end{table}

\textbf{Effect of the future time $\tau_f$.} Table~\ref{tab:ablation-future-time} shows the results. We see an expected trend that the task becomes more difficult as $\tau_f$ increases. However, at a very high $\tau_f$, the interaction location prediction becomes an easier task since the person has navigated to a large part of the environment, thus making majority of the locations as ground truth. Thus, we choose $\tau_f=180s$ as a challenging version of the future interaction location prediction.

\begin{table}[h!]\footnotesize
    \centering
    \begin{tabular}{cc}
        Location prediction & Pose prediction \\ %
        \begin{tabular}{cccc}
             $60s$ & $120s$ & $180s$ & $600s$ \\ \hline
            \textbf{22.6} & 21.4 & 21.2 & 21.4 \\ \hline
        \end{tabular}
        &
        \begin{tabular}{cccc}
             $60s$ & $120s$ & $180s$ & $600s$ \\ \hline
            \textbf{207} & 212 & 215 & 220 \\ \hline
        \end{tabular}
    \end{tabular}
    \caption{Effect of $\tau_f$ on the performance.}
    \label{tab:ablation-future-time}
\end{table}

\textbf{Effect of the learning rate.} Table~\ref{tab:ablation-lr} shows the results. We see that the model performs the best with a learning rate of $5\times 10^{-5}$ for interaction location prediction, and $5 \times 10^{-6}$ for future pose prediction. This same parameter is chosen for all testing, as mentioned in Sec.~\ref{sec:implementation}.

\begin{table}[h!]\footnotesize
    \centering
    \begin{tabular}{cc}
        Location prediction & Pose prediction \\ %
        \begin{tabular}{ccc}
             $5.10^{-6}$ & $5.10^{-5}$ & $5.10^{-4}$ \\ \hline
            20.6 & \textbf{21.2} & 19.6 \\ \hline
        \end{tabular}
        &
        \begin{tabular}{ccc}
             $5.10^{-6}$ & $5.10^{-5}$ & $5.10^{-4}$\\ \hline
            \textbf{215} & 220 & 226 \\ \hline
        \end{tabular}
    \end{tabular}
    \caption{Effect of learning rate on the performance.}
    \label{tab:ablation-lr}
\end{table}

\textbf{Effect of the encoder model size.} We use a simple transformer encoder $\mathcal{L}$ for encoding the environment context (Sec. \ref{sec:model}). We experiment with varying number of transformer layers. We experiment with $2, 4$ and $6$ layers. Table~\ref{tab:ablation-model-size} shows the results. We observe that the number increases with the number of layers. This suggests that the performance can be further improved, with a larger transformer size. We do not experiment beyond $6$ due to hardware constraints.

\begin{table}[h!]\footnotesize
    \centering
    \begin{tabular}{cc}
        Location prediction & Pose prediction \\ %
        \begin{tabular}{ccc}
             $2$ & $4$ & $6$ \\ \hline
            20.2 & 20.9 & \textbf{21.2} \\ \hline
        \end{tabular}
        &
        \begin{tabular}{ccc}
             $2$ & $4$ & $6$ \\ \hline
            228 & 222 & \textbf{215} \\ \hline
        \end{tabular}
    \end{tabular}
    \caption{Effect of the model size on the performance. We vary the number of transformer layers.}
    \label{tab:ablation-model-size}
\end{table}

\section{Limitations}

As discussed in Sec. \ref{sec:method}, our current method assumed one actor per video. The model design cannot explicitly handle multi-person scenarios. We will handle multi-person scenarios in the future. Nevertheless, the single-actor problem is still challenging with scope for improvement. We also assume a static point cloud when creating the dataset, while in practice, the object location can change with time. It is possible to use 3D information only from the last time segment for improving the spatial input to the model, we do not consider this case for the ease of the I/O. Note that this simplification does not affect the curated dataset quality, since we use narrations from Ego-Exo4D \cite{egoexo4d} as an additional signal. Finally, we use state-of-the-art methods Detic \cite{detic}, WHAM \cite{wham} and Llama 3.1 \cite{llama-herd} for creating the dataset, which are prone to errors. Any future improvement in these domains will further strengthen our dataset quality and the resulting trained model.

\clearpage